\documentclass[11pt,a4paper]{article}
\usepackage[hyperref]{naaclhlt2018}
\usepackage{times}
\usepackage{latexsym}

\usepackage{url}

\usepackage{epsfig}
\usepackage{graphicx}
\usepackage{amsmath}
\usepackage{amssymb}

\usepackage{epstopdf}
\usepackage{enumitem}
\usepackage{amsmath,amsfonts}
\usepackage{amssymb,amsopn}
\usepackage{bm} % bold symbol

\usepackage{amssymb}
\usepackage{amsmath,amsfonts}
\usepackage{amsthm,amsopn}

\usepackage{bm} % bold symbol

\newcommand{\QU}{QoU\xspace}
\newcommand{\IU}{IoU\xspace}
\newcommand{\AU}{Neutrality}
\newcommand{\AUflat}{Neutrality}
\newcommand{\QUflat}{Question only Unresolvable (QoU)}
\newcommand{\IUflat}{Image only Unresolvable (IoU)}

\newcommand{\mat}[1]{\boldsymbol{#1}} % matrix
\newcommand{\ProbOpr}[1]{\mathbb{#1}}
\newcommand{\expect}[2]{%
\ifthenelse{\equal{#2}{}}{\ProbOpr{E}_{#1}}
{\ifthenelse{\equal{#1}{}}{\ProbOpr{E}\left[#2\right]}{\ProbOpr{E}_{#1}\left[#2\right]}}} % Expectation: syntax: E{1}{2} = E_1[2], E{}{2}=E[2], E{1}{} = E_1
\newcommand{\var}[2]{%
\ifthenelse{\equal{#2}{}}{\ProbOpr{VAR}_{#1}}
{\ifthenelse{\equal{#1}{}}{\ProbOpr{VAR}\left[#2\right]}{\ProbOpr{VAR}_{#1}\left[#2\right]}}} % Expectation: syntax: V{1}{2} = V_1[2], V{}{2}=V[2], V{1}{} = V_1
\newcommand{\mU}{\mat{U}}

\newcommand{\mW}{\mat{W}}

\newcommand{\eat}[1]{}

\usepackage{multirow}
\usepackage{array}
\usepackage{dsfont}
\usepackage{boldline}
\usepackage{dblfloatfix}
\usepackage{setspace}

\usepackage{xspace}

\aclfinalcopy

\newcommand*{\ie}{i.e.\@\xspace}

\begin{document}

%%%%%%%%% TITLE
\title{Being Negative but Constructively: \\
        Lessons Learnt from Creating Better Visual Question Answering Datasets}

\author{Wei-Lun Chao$^*$, Hexiang Hu\thanks{\hspace{4pt}Equal contributions} , Fei Sha\\
University of Southern California\\
Los Angeles, California, USA\\
{\tt\small  weilunchao760414@gmail.com, hexiang.frank.hu@gmail.com, feisha@usc.edu}
}

\maketitle

% !TEX root = main.tex
\begin{abstract}
Visual question answering (Visual QA) has attracted a lot of attention lately, seen essentially as a form of (visual) Turing test that artificial intelligence should strive to achieve. In this paper, we study a crucial component of this task: how can we design good datasets for the task?  We focus on the design of multiple-choice based datasets where the learner has to select the right answer from a set of candidate ones including the target (\ie the correct one) and the decoys (\ie the incorrect ones). Through careful analysis of the results attained by state-of-the-art learning models and human annotators on existing datasets, we show that the design of the decoy answers has a significant impact on how and what the learning models learn from the datasets. In particular, the resulting learner can ignore the visual information, the question, or both while still doing well on the task. Inspired by this, we propose automatic procedures  to remedy such design deficiencies. We apply the procedures to re-construct decoy answers for two popular Visual QA datasets as well as to create a new Visual QA dataset from the Visual Genome project, resulting in the largest dataset for this task. Extensive empirical studies show that the design deficiencies have been alleviated in the remedied datasets and the performance on them is likely a more faithful indicator of the difference among learning models. The datasets are released and publicly available via \url{http://www.teds.usc.edu/website_vqa/}.
\end{abstract}
% !TEX root = main.tex
\begin{figure}[t]
\centering
\includegraphics[width=0.49\textwidth]{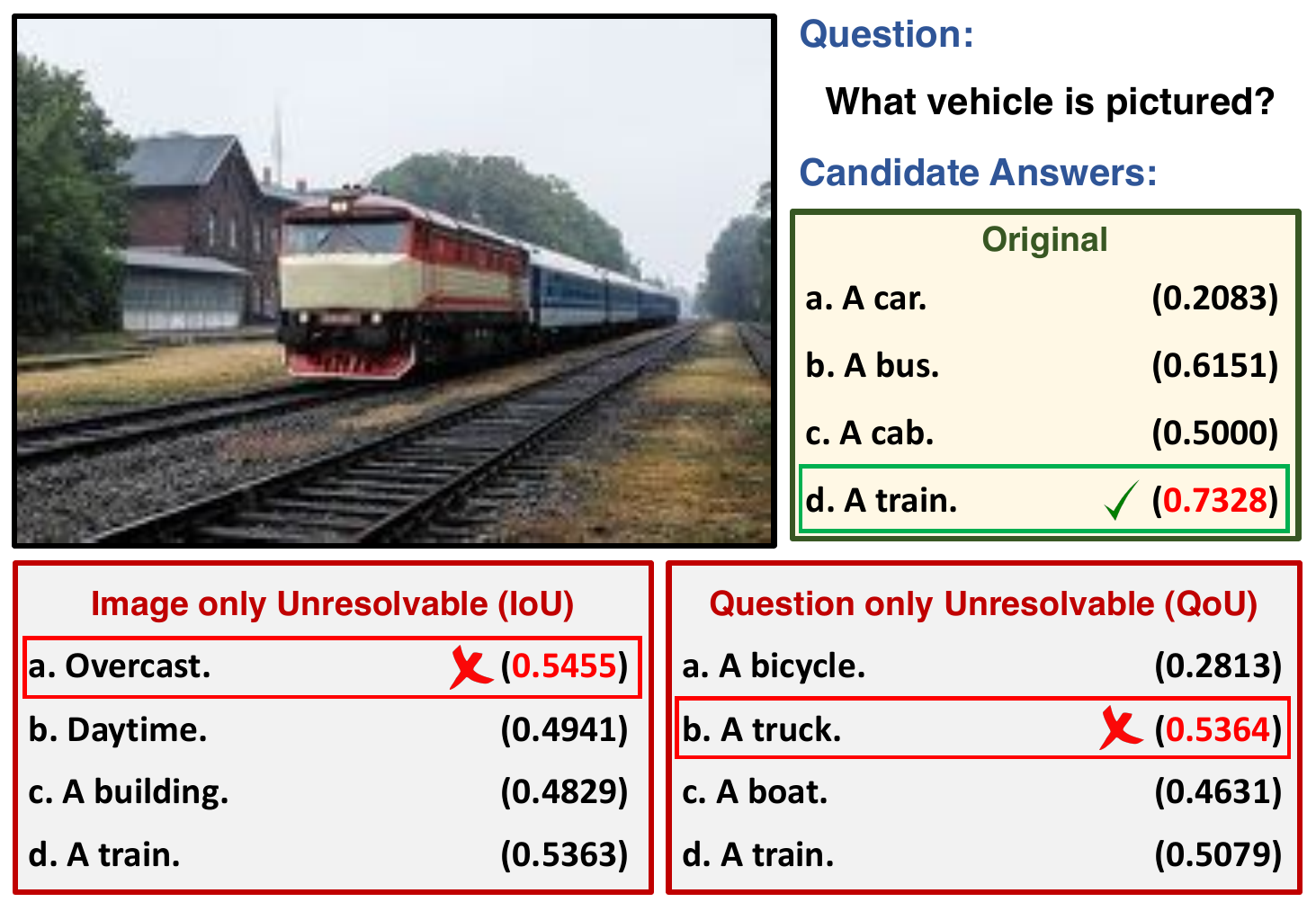}
\caption{An illustration of how the shortcuts in the Visual7W dataset~\cite{zhu2016Visual7W} should be remedied. In the original dataset, the correct answer ``A train'' is easily selected by a machine as it is far often used as the correct answer than the other decoy (negative) answers. (The numbers in the brackets are probability scores computed using eq. (\ref{e_prob})). Our two procedures --- \QU and \IU (cf. Sect.~\ref{S_method}) --- create alternative decoys such that both the correct answer and the decoys are highly likely by examining either the image or the question \textbf{alone}. In these cases, machines make mistakes unless they consider all information \textbf{together}. Thus, the alternative decoys suggested our procedures are better designed to gauge how well a learning algorithm can understand all information equally well.}
\label{fConcept}
\vspace{-10pt}
\end{figure}

\section{Introduction}

Multimodal information processing tasks such as image captioning~\cite{farhadi2010every,ordonez2011im2text,xu2015show} and visual question answering (Visual QA)~\cite{antol2015vqa} have gained a lot of attention recently. A number of significant advances in learning algorithms have been made, along with the development of nearly two dozens of datasets in this very active research domain. Among those datasets, popular ones include MSCOCO~\cite{lin2014mscoco,chen2015microsoft}, Visual Genome~\cite{krishna2016vg}, VQA~\cite{antol2015vqa}, and several others. The overarching objective is that a learning machine needs to go beyond understanding different modalities of information separately (such as image recognition alone) and to learn how to correlate them in order to perform well on those tasks. 

To evaluate the progress on those complex and more AI-like tasks is however a challenging topic.  For tasks involving language generation, developing an automatic evaluation metric is itself an open problem~\cite{anderson2016spice,kilickaya2016re,liu2016not,kafle2016visual}. Thus, many efforts have concentrated on tasks such as \emph{multiple-choice} Visual QA~\cite{antol2015vqa,zhu2016Visual7W,jabri2016revisiting} or selecting the best caption~\cite{hodosh2013framing,hodosh2016focused,ding2016understanding,lin2016leveraging}, where the selection accuracy is a natural evaluation metric. 

In this paper, we study how to design high-quality multiple choices for the Visual QA task.    In this task,  the machine (or the human annotator) is presented with an image, a question and a list of candidate answers. The goal is to select the correct answer through a consistent understanding of the image, the question and each of the candidate answers. As in any multiple-choice based  tests (such as GRE), designing what should be presented as negative answers --- we refer them as \emph{decoys} --- is as important as deciding the questions to ask.  We all have had the experience of exploiting the elimination strategy: \emph{This question is easy --- none of the three answers could be right so the remaining one must be correct!}

While a clever strategy for taking exams, such ``shortcuts'' prevent us from studying faithfully how different learning algorithms comprehend the meanings in images and languages (e.g., the quality of the embeddings of both images and languages in a semantic space). It has been noted that  machines can achieve very high accuracies of selecting the correct answer without the visual input (\ie, the image), the question, or both~\cite{jabri2016revisiting,antol2015vqa}. Clearly, the learning algorithms have overfit on incidental statistics in the datasets. For instance, if the decoy answers have rarely been used as the correct answers (to any questions), then the machine can rule out a decoy answer with a binary classifier that determines whether the answers are in the set of the correct answers --- note that this classifier does not need to examine the image and it just needs to memorize the list of the correct answers in the training dataset. See Fig.~\ref{fConcept} for an example, and Sect.~\ref{S_diagnosis} for more and detailed analysis.

We focus on minimizing the impacts of exploiting such shortcuts. We suggest a set of principles for creating decoy answers.  In light of the amount of human efforts in curating existing datasets for the Visual QA task, we propose two procedures that revise those datasets such that the decoy answers are better designed.  In contrast to some earlier works, the procedures are fully automatic and do not incur additional human annotator efforts. We apply the procedures to revise both Visual7W~\cite{zhu2016Visual7W} and VQA~\cite{antol2015vqa}. Additionally, we create new multiple-choice based datasets from COCOQA~\cite{ren2015exploring} and the recently released VQA2~\cite{goyal2016making} and Visual Genome datasets~\cite{krishna2016vg}. The one based on Visual Genome becomes the largest multiple-choice dataset for the Visual QA task, with more than one million image-question-candidate answers triplets.  

We conduct  extensive empirical and human studies to demonstrate the effectiveness of our procedures in creating high-quality datasets for the Visual QA task. In particular, we show that machines need to use all three information (image, questions and answers) to perform well --- any missing information induces a large drop in performance. Furthermore, we show that humans dominate machines in the task. However, given the revised datasets are likely reflecting the true gap between the human and the machine understanding of multimodal information, we expect that advances in learning algorithms likely focus more on the task itself instead of overfitting to the idiosyncrasies in the datasets.

The rest of the paper is organized as follows. In Sect.~\ref{S_related}, we describe related work. In Sect.~\ref{S_diagnosis}, we analyze and discuss the design deficiencies in existing datasets. In Sect.~\ref{S_method}, we describe our automatic procedures for remedying those deficiencies. In Sect.~\ref{S_exp} we conduct experiments and analysis. We conclude the paper in Sect.~\ref{S_disc}.

% !TEX root = main.tex
\section{Related Work}
\label{S_related}

Wu et al. \shortcite{wu2016visual} and Kafle and Kanan \shortcite{kafle2016visual} provide recent overviews of the status quo of the Visual QA task. There are about two dozens of datasets for the task. Most of them use real-world images, while some are based on synthetic ones. Usually, for each image, multiple questions and their corresponding answers are generated. This can be achieved either by human annotators, or with an automatic procedure that uses captions or question templates and detailed image annotations. We concentrate on
3 datasets: VQA~\cite{antol2015vqa}, Visual7W~\cite{zhu2016Visual7W}, and Visual Genome~\cite{krishna2016vg}. All of them use images from MSCOCO~\cite{lin2014mscoco}.

Besides the pairs of questions and correct answers, VQA, Visual7W, and visual Madlibs~\cite{yu2015visual} provide decoy answers for each pair so that the task can be evaluated in multiple-choice selection accuracy. \emph{What decoy answers to use} is the focus of our work.

In VQA, the decoys consist of human-generated plausible answers as well as high-frequency and random answers from the datasets. In Visual7W, the decoys are all human-generated plausible ones. Note that, humans generate those decoys by \emph{only looking at the questions and the correct answers but \textbf{not} the images}. Thus, the decoys might be unrelated to the corresponding images. A learning algorithm can potentially examine the image alone and be able to identify the correct answer.

In visual Madlibs, the questions are generated with a limited set of question templates and the detailed annotations (e.g., objects) of the images. Thus, similarly, a learning model can examine the image alone and deduce the correct answer. 

We propose automatic procedures to revise VQA and Visual7W (and to create new datasets based on COCOQA~\cite{ren2015exploring}, VQA2~\cite{goyal2016making}, and Visual Genome) such that the decoy generation is carefully orchestrated to prevent learning algorithms from exploiting the shortcuts in the datasets by overfitting on incidental statistics. In particular, our design goal is that a learning machine needs to understand all the 3 components of an image-question-candidate answers triplet in order to make the right choice --- ignoring either one or two components will result in drastic degradation in performance.

Our work is inspired by the experiments in~\cite{jabri2016revisiting} where they observe that machines without looking at images or questions can still perform well on the Visual QA task.  Others have also reported similar issues~\cite{goyal2016making,zhang2016yin,johnson2016clevr,agrawal2016analyzing,Kafle_2017_ICCV, agrawal2018don}, though not in the multiple-choice setting. Our work extends theirs by providing more detailed analysis \emph{as well as automatic procedures} to remedy those design deficiencies.

Besides Visual QA, VisDial~\cite{das2016visual} and Ding et al. \shortcite{ding2016understanding} also propose automatic ways to generate decoys for the tasks of multiple-choice visual captioning and dialog, respectively.

Recently, Lin and Parikh \shortcite{lin2017active} study active learning for Visual QA: \ie, how to select informative image-question pairs (for acquiring annotations) or image-question-answer triplets for machines to ``learn'' from. On the other hand, our work further focuses on designing better datasets for ``evaluating'' a machine.

% !TEX root = main.tex

\section{Analysis of Decoy Answers' Effects}
\label{S_diagnosis}

In this section, we examine in detail the dataset Visual7W~\cite{zhu2016Visual7W}, a popular choice for the Visual QA task. We demonstrate how the deficiencies in designing decoy questions impact the performance of learning algorithms.

In multiple-choice Visual QA datasets, a training or test example is a triplet that consists of an image I, a question Q, and a candidate answer set A. The set A contains a target T (the correct answer)  and $K$  decoys  (incorrect answers) denoted by D. An IQA triplet is thus $\{\text{I}, \text{Q},\text{A}=\{\text{T},\text{D}_1,\cdots,\text{D}_K\}\}$. We use C to denote either the target or a decoy.

 \subsection{Visual QA models}
\label{S_diagnosis_model}

We investigate how well a learning algorithm can perform when supplied with different modalities of information.  We concentrate on the one hidden-layer MLP model proposed in \cite{jabri2016revisiting}, which has achieved state-of-the-art results on the dataset Visual7W. The model computes a scoring function $f(c, i)$
\begin{align}
f(c,i)=\sigma(\mU \max(0,\mW g(c,i)) + b)
\label{evqamodel}
\end{align}
over a candidate answer $c$ and the multimodal information $i$, where $g$ is the joint feature of $(c, i)$ and $\sigma(x)=1/(1+\exp(-x))$. The information $i$ can be null, the image (I) alone, the question (Q) alone, or the combination of both (I+Q).

Given an IQA triplet, we use the penultimate layer of ResNet-200~\cite{he2016deep} as visual features to represent I and the average \textsc{word2vec} embeddings~\cite{mikolov2013distributed} as text features to represent Q and C. To form the joint feature $g(c,i)$, we just concatenate the features together. The candidate $c\in\text{A}$ that has the highest $f(c,i)$ score in prediction is selected as the model output.

We use the standard training, validation, and test splits of Visual7W, where each contains 69,817, 28,020, and 42,031 examples respectively. Each question has 4 candidate answers. The parameters of $f(c,i)$ are learned by minimizing the binary logistic loss of predicting whether or not a candidate $c$ is the target of an IQA triplet. Details are in Sect.~\ref{S_exp} and the Supplementary Material.

\begin{table}[t]
\vskip 3pt
\small
\centering
\begin{tabular}{l|c|c}
Information used & Machine & Human\\
\hline
random    & 25.0  & 25.0 \\
A         & 52.9  & - \\
I + A     & 62.4  & 75.3 \\
Q + A     & 58.2  & 36.4 \\
I + Q + A & 65.7  & 88.4 \\
\hline
\end{tabular}
\vskip -3pt
\caption{Accuracy of selecting the right answers out of 4 choices (\%) on the Visual QA task on Visual7W.}
\label{t_flaw_v7w}
\vspace{-5pt}
\end{table}

\subsection{Analysis results}

\paragraph{Machines find shortcuts} Table~\ref{t_flaw_v7w} summarizes the performance of the learning models, together with the human studies we performed on a subset of 1,000 triplets (c.f. Sect.~\ref{S_exp} for details). There are a few interesting observations.

First, in the row of ``A'' where only the candidate answers (and whether they are right or wrong) are used to train a learning model, the model performs significantly better than random guessing and humans (52.9\% vs. 25\%) --- humans will deem each of the answers equally likely \emph{without} looking at both the image and the question! Note that in this case, the information $i$ in eq.~(\ref{evqamodel}) contains nothing. The model learns the specific statistics of the candidate answers in the dataset and exploits those.
Adding the information about the image (\ie, the row of ``I+A''), the machine improves significantly and gets close to the performance when all information is used (62.4\% vs. 65.7\%).  There is a weaker correlation between the question and the answers as ``Q+A'' improves over ``A'' only modestly. This is expected. In the Visual7W dataset, the decoys are generated by human annotators as plausible answers to the questions without being shown the images --- thus, many decoy answers do not have visual groundings. For instance, a question of ``what animal is running?'' elicits equally likely answers such as ``dog'', ``tiger'', ``lion'', or ``cat'', while an image of a dog running in the park will immediately rule out all 3 but the ``dog'', see Fig.~\ref{fConcept} for a similar example.  Thus, the performance of ``I+A'' implies that many IQA triplets can be solved by object, attribute or concept detection on the image, without understanding the questions. This is indeed the case also for humans --- humans can achieve 75.3\% by considering ``I+A'' and not ``Q''.  Note that the difference between machine and human on ``I+A'' are likely due to their difference in understanding visual information.

Note that human improves significantly from ``I+A'' to ``I+Q+A'' with ``Q'' added, while the machine does so only marginally. The difference can be attributed to the difference in understanding the question and correlating with the answers between the two. Since each image corresponds to multiple questions or have multiple objects, solely relying on the image itself will not work well in principle. Such difference clearly indicates that in the Visual QA model, the language component is weak as the model cannot fully exploit the information in ``Q'', making a smaller relative improvement 5.3\% (from 62.4\% to 65.7\%) where humans improved relatively 17.4\%.

\paragraph{Shortcuts are due to design deficiencies} We probe deeper on how the decoy answers have impacted the performance of learning models.

As explained above, the decoys are drawn from all plausible answers to a question, irrespective of whether they are visually grounded or not. We have also discovered that the targets (\ie, correct answers) are infrequently used as decoys. 

Specifically, among the 69,817 training samples, there are 19,503 unique correct answers and each one of them is used about 3.6 times as correct answers to a question.  However, among all the $69,817 \times 3 \approx 210K$ decoys, each correct answer appears 7.2 times on average, far below a chance level of 10.7 times ($210K \div  19,503 \approx  10.7$). This disparity exists in the test samples too. Consequently, the following rule, computing each answer's likelihood of being correct,
\begin{align}
& P(\text{correct}|\text{C})= \nonumber\\
&
\begin{cases}
 0.5, & \hspace{-7.4em}\text{if C is never seen in training,}\\
\frac{\text{\# times C as target}}{\text{\# times C as target} + (\text{\# times C as decoys})/K}, & \text{otherwise,}
\end{cases}
\label{e_prob}
\end{align}
should perform well.  Essentially, it measures how unbiased C is used as the target and the decoys. Indeed, it attains an accuracy of 48.73\% on the test data, far better than the random guess and is close to the learning model using the answers' information only (the ``A'' row in Table~\ref{t_flaw_v7w}).

\paragraph{Good rules for designing decoys} Based on our analysis, we  summarize the following guidance rules to design decoys:  (1) \textbf{\QUflat}. The decoys need to be equally plausible to the question. Otherwise, machines can rely on the correlation between the question and candidate answers to tell the target from decoys, even without the images. Note that this is a principle that is being followed by most datasets. (2) \textbf{\AUflat}. The decoys answers should be equally likely used as the correct answers. (3) \textbf{\IUflat}. The decoys need to be plausible to the image. That is, they should appear in the image, or there exist questions so that the decoys can be treated as targets to the image. Otherwise, Visual QA can be resolved by objects, attributes, or concepts detection in images, even without the questions.

Ideally, each decoy in an IQA triplet should meet the three principles. \textbf{\AUflat} is comparably easier to achieve by \emph{reusing terms in the whole set of targets as decoys.} On the contrary, a decoy may hardly meet \textbf{\QU} and \textbf{\IU} simultaneously\footnote{E.g., in Fig~\ref{fConcept}, for the question ``What vehicle is pictured?'', the only answer that meets both principles is ``train'', which is the correct answer instead of being a decoy.}. However, as long as all decoys of an IQA triplet meet \textbf{\AUflat} and some meet \textbf{\QU} and others meet \textbf{\IU}, the triplet as a whole still achieves the three principles --- a machine ignoring either images or questions will likely perform poorly.

% !TEX root = main.tex
\section{Creating Better Visual QA Datasets}
\label{S_method}

In this section, we describe our approaches of remedying design deficiencies in the existing datasets for the Visual QA task. We introduce two automatic and widely-applicable procedures to create new decoys that can prevent learning models from exploiting incident statistics in the datasets.

\subsection{Methods}
\label{s_Method}
\paragraph{Main ideas} Our procedures operate on a dataset that already contains image-question-target (IQT) triplets, \ie, we do not assume it has decoys already. For instance, we have used our procedures to create a multiple-choice dataset from the Visual Genome dataset which has no decoy.  We assume that each image in the dataset is coupled with ``multiple" QT pairs, which is the case in nearly all the existing datasets. Given an IQT triplet (I, Q, T), we create  two sets of decoy answers.
\begin{itemize}[leftmargin=2\labelsep]
\item \textbf{\QU-decoys}. We search among all other triplets that have similar questions to Q. The \textbf{targets} of those triplets are then collected as the decoys for T. As the targets to similar questions are likely plausible for the question Q, \QU-decoys likely follow the rules of  \textbf{\AUflat} and \textbf{\QUflat}. We compute  the average \textsc{word2vec}~\cite{mikolov2013distributed} to represent a question, and use the cosine similarity to measure the  similarity between questions.

\item \textbf{\IU-decoys}. We collect the \textbf{targets} from other triplets of the \emph{same} image to be the decoys for T. The resulting decoys thus definitely follow the rules of \textbf{\AUflat} and \textbf{\IUflat}.
\end{itemize}
We then combine the triplet (I, Q, T) with \QU-decoys and \IU-decoys to form an IQA triplet as a training or test sample.

\paragraph{Resolving ambiguous decoys}
One potential drawback of automatically selected decoys is that they may be semantically similar, ambiguous, or rephrased terms to the target~\cite{zhu2016Visual7W}. We utilize two filtering steps to alleviate it. First, we perform string matching between a decoy and the target, deleting those decoys that contain or are covered by the target (e.g., ``daytime" vs. ``during the daytime" and ``ponytail" vs. ``pony tail").

Secondly, we utilize the WordNet hierarchy and the Wu-Palmer (WUP) score~\cite{wu1994verbs} to eliminate semantically similar decoys. The WUP score measures how similar two \emph{word senses} are (in the range of $[0, 1]$), based on the depth of them in the taxonomy and that of their least common subsumer. We compute the similarity of two strings according to the WUP scores in a similar manner to~\cite{malinowski2014multi}, in which the WUP score is used to evaluate Visual QA performance. We eliminate decoys that have higher WUP-based similarity to the target. We use the NLTK toolkit~\cite{bird2009natural} to compute the similarity. See the Supplementary Material for more details.

\paragraph{Other details}
For \QU-decoys, we sort and keep for each triplet the top $N$ (e.g., 10,000) similar triplets from the entire dataset according to the question similarity. Then for each triplet, we compute the WUP-based similarity of each potential decoy to the target successively, and accept those with similarity below 0.9 until we have $K$ decoys. We choose 0.9 according to~\cite{malinowski2014multi}. We also perform such a check among selected decoys to ensure they are not very similar to each other. For \IU-decoys, the potential decoys are sorted randomly. The WUP-based similarity with a threshold of 0.9 is then applied to remove ambiguous decoys.

\subsection{Comparison to other datasets}

Several authors have noticed the design deficiencies in the existing databases and have proposed ``fixes''~\cite{antol2015vqa,yu2015visual,zhu2016Visual7W,das2016visual}. No dataset has used a procedure to generate \IU-decoys. We empirically show that how the \IU-decoys significantly remedy the design deficiencies in the datasets.

Several previous efforts have generated decoys that are similar in spirit to our \textbf{\QU}-decoys. Yu et al.~\shortcite{yu2015visual}, Das et al.~\shortcite{das2016visual}, and Ding et al.~\shortcite{ding2016understanding} automatically find decoys from similar questions or captions based on question templates and annotated objects, tri-grams and \textsc{GloVe} embeddings~\cite{pennington2014glove}, and paragraph vectors~\cite{le2014distributed} and linguistic surface similarity, respectively. The later two are for different tasks from Visual QA, and only Ding et al.~\shortcite{ding2016understanding} consider removing semantically ambiguous decoys like ours. Antol et al.~\shortcite{antol2015vqa} and Zhu et al.~\shortcite{zhu2016Visual7W} ask humans to create decoys, given the questions and targets. As shown earlier,  such decoys may disobey the rule of \textbf{\AU}.

Goyal et al.~\shortcite{goyal2016making} augment the VQA dataset \cite{antol2015vqa} (by human efforts) with additional IQT triplets to eliminate the shortcuts (language prior) in the open-ended setting. Their effort is complementary to ours on the multiple-choice setting. Note that an extended task of Visual QA, visual dialog~\cite{das2016visual}, also adopts the latter setting.

% !TEX root = main.tex
\section{Empirical Studies}
\label{S_exp}

\subsection{Dataset}
% !TEX root = main.tex
We examine our automatic procedures for creating decoys on five datasets. Table~\ref{t_dataset} summarizes the characteristics of the three datasets we focus on.

\paragraph{VQA Real~\cite{antol2015vqa}} The dataset uses images from MSCOCO~\cite{lin2014mscoco} under the same training/validation/testing splits to construct IQA triplets. Totally 614,163 IQA triplets are generated for 204,721 images. 
Each question has 18 candidate answers: in general 3 decoys are human-generated, 4 are randomly sampled, and 10 are randomly sampled frequent-occurring targets. \emph{As the test set does not indicate the targets, our studies focus on the training and validation sets.}

\paragraph{Visual7W Telling (Visual7W)~\cite{zhu2016Visual7W}} The dataset uses 47,300 images from MSCOCO \cite{lin2014mscoco} and contains 139,868 IQA triplets. Each has 3 decoys generated by humans.

\paragraph{Visual Genome (VG) \cite{krishna2016vg}} The dataset uses 101,174 images from MSCOCO \cite{lin2014mscoco} and contains 1,445,322 IQT triplets. No decoys are provided. Human annotators are asked to write diverse pairs of questions and answers freely about an image or with respect to some regions of it. On average an image is coupled with 14 question-answer pairs. We divide the dataset into non-overlapping 50\%/20\%/30\% for training/validation/testing. Additionally, we partition such that each portion is a ``superset'' of the corresponding one in Visual7W, respectively.

\paragraph{VQA2~\cite{goyal2016making} and COCOQA~\cite{ren2015exploring}} We describe the datasets and experimental results in the Supplementary Material.

\paragraph{Creating decoys}
We create 3 \QU-decoys and 3 \IU-decoys for every IQT triplet in each dataset, following the steps in Sect.~\ref{s_Method}. In the cases that we cannot find 3 decoys, we include random ones from the original set of decoys for VQA and Visual7W; for other datasets, we randomly include those from the top 10 frequently-occurring targets. 

\begin{table}[t]
\vskip 3pt
\small
\tabcolsep 2.5pt
\centering
\begin{tabular}{c|rrr|rrr|c}
{Dataset}&  \multicolumn{3}{c|}{\# of Images} &  \multicolumn{3}{c|}{\# of triplets} & {\# of decoys}\\ \cline{2-7}
{Name} & train & val & test  & train & val & test & per triplet\\
\hline
{VQA} & 83k & 41k & 81k & 248k & 121k & 244k & 17\\ 
{Visual7W} & 14k & 5k  & 8k & 69k & 28k & 42k & 3\\ 
{VG}  & 49k & 19k & 29k & 727k & 283k & 433k & - \\ 
\hline
\end{tabular}
\vskip -3pt
\caption{Summary of Visual QA datasets.}
\label{t_dataset}
\vspace{-5pt}
\end{table}

\subsection{Setup}
% !TEX root = main.tex
\paragraph{Visual QA models}
We utilize the MLP models mentioned in Sect.~\ref{S_diagnosis} for all the experiments. We denote \textbf{MLP-A}, \textbf{MLP-QA}, \textbf{MLP-IA}, \textbf{MLP-IQA} as the models using A (Answers only), Q+A (Question plus Answers), I+A (Image plus Answers), and I+Q+A (Image, Question and Answers) for multimodal information, respectively. The hidden-layer has 8,192 neurons. We use a 200-layer ResNet~\cite{he2016deep} to compute visual features which are  2,048-dimensional. The ResNet is pre-trained on ImageNet~\cite{russakovsky15imagenet}. The \textsc{word2vec} feature~\cite{mikolov2013distributed} for questions and answers are 300-dimensional, pre-trained on Google News\footnote{We experiment on using different features in the Supplementary Material.}. The parameters of the MLP models are learned by minimizing the binary logistic loss of predicting whether or not a candidate answer is the target of the corresponding IQA triplet. Please see the Supplementary Material for details on optimization. 

We further experiment with a variant of the spatial memory network (denoted as \textbf{Attention}) \cite{xu2016ask} and the \textbf{HieCoAtt} model \cite{lu2016hierarchical} adjusted for the multiple-choice setting. Both models utilize the attention mechanism. Details are listed in the Supplementary Material.

\paragraph{Evaluation metric}
For VQA and VQA2, we follow their protocols by comparing the picked answer to 10 human-generated targets. The accuracy is computed based on the number of exactly matched targets (divided by 3 and clipped at 1). For others, we compute the accuracy of picking the target from multiple choices. 

\paragraph{Decoy sets to compare}

For each dataset, we derive several variants: (1) \textbf{Orig}: the original decoys from the datasets, (2) \textbf{\QU}: \textbf{Orig} replaced with ones selected by our  \QU-decoys generating procedure,  (3) \textbf{\IU}:  \textbf{Orig} replaced with ones selected by our \IU-decoys generating procedure, (4) \textbf{\QU+\IU}:  \textbf{Orig} replaced with ones  combining \textbf{\QU} and \textbf{\IU}, (5) \textbf{All}: combining \textbf{Orig}, \textbf{\QU}, and \textbf{\IU}.

\paragraph{User studies} Automatic decoy generation may lead to ambiguous decoys as mentioned in Sect.~\ref{S_method} and \cite{zhu2016Visual7W}. We conduct a user study via Amazon Mechanic Turk (AMT) to test humans' performance on the datasets after they are remedied by our automatic procedures. We select 1,000 IQA triplets from each dataset. Each triplet is answered by three workers and in total 169 workers get involved. The total cost is \$215 --- the rate for every 20 triplets is \$0.25. We report the average human performance and compare it to the learning models'. See the Supplementary Material for more details.

\subsection{Results}
% !TEX root = main.tex
The performances of learning models and humans on the 3 datasets are reported in Table~\ref{v7w-1}, \ref{vqa-1}, and \ref{VG-1}\footnote{We note that in Table~\ref{v7w-1}, the 4.3\% drop of the human performance on \IU+\QU, compared to Orig, is likely due to that \IU+\QU has more candidates (7 per question). Besides, the human performance on qaVG cannot be directly compared to that on the other datasets, since the questions on qaVG tend to focus on local image regions and are considered harder.}.

\begin{table}
	\vskip 3pt
	\small
	\centering
	\tabcolsep 5pt
	\begin{tabular}{l|c|c|c|c|c}
		Method & Orig & \IU & \QU & \IU+\QU & All \\
		\hline
		MLP-A   & 52.9 & 27.0 & 34.1 & 17.7 & 15.6 \\
		MLP-IA  & 62.4 & 27.3 & 55.0 & 23.6 & 22.2 \\
		MLP-QA  & 58.2 & 84.1 & 40.7 & 37.8 & 31.9 \\
		MLP-IQA & 65.7 & 84.1 & 57.6 & 52.0 & 45.1 \\
		\hline
		HieCoAtt$*$ & 63.9 & - & - & 51.5 & - \\
		Attntion$*$ &  65.9 & - & - & 52.8 & - \\
		\hline
		Human  & 88.4 & - & - & 84.1 & - \\
		\hline
		Random & 25.0 & 25.0 & 25.0 & 14.3 & 10.0 \\
	\end{tabular}
	{\\$*$: based on our implementation or modification}
	\vskip -3 pt
	\caption{Test accuracy (\%) on Visual7W.}
	\label{v7w-1}
	\vskip -5pt
\end{table}

\paragraph{Effectiveness of new decoys} A better set of decoys will force learning models to integrate all 3 pieces of information --- images, questions and answers --- to make the correct selection from multiple-choices. In particular, they should prevent learning algorithms from exploiting shortcuts such that partial information is sufficient for performing well on the Visual QA task.

Table~\ref{v7w-1} clearly indicates that those goals have been achieved. With the Orig decoys, the relatively small gain from MLP-IA to MLP-IQA suggests that the question information can be ignored to attain good performance. However, with the \IU-decoys which require questions to help to resolve (as image itself is inadequate to resolve), the gain is substantial (from 27.3\% to 84.1\%). Likewise, with the \QU-decoys (question itself is not adequate to resolve), including images information improves from 40.7\% (MLP-QA) substantially to 57.6\% (MLP-IQA). Note that with the Orig decoys, this gain is smaller (58.2\% vs. 65.7\%).

It is expected that MLP-IA matches better \QU-decoys but not \IU-decoys, and MLP-QA is the other way around.  Thus it is natural to combine these two decoys. What is particularly appealing is that MLP-IQA improves noticeably over models learned with partial information on the combined \IU+\QU -decoys (and ``All'' decoys\footnote{We note that the decoys in Orig are not trivial, which can be seen from the gap between All and \IU+\QU. Our main concern on Orig is that for those questions that machines can accurately answer, they mostly rely on only partial information. This will thus hinder designing machines to fully comprehend and reason from multimodal information. We further experiment on random decoys, which can achieve \textbf{\AUflat} but not the other two principles, to demonstrate the effectiveness of our methods in the Supplementary Material.}).  Furthermore, using answer information only (MLP-A) attains about the chance-level accuracy.

On the VQA dataset (Table~\ref{vqa-1}), the same observations hold, though to a lesser degree. On any of the \IU\ or \QU\ columns, we observe substantial gains when the complementary information is added to the model (such as MLP-IA to MLP-IQA). All these improvements are much more visible than those observed on the original decoy sets.

Combining both Table~\ref{v7w-1} and \ref{vqa-1}, we notice that the improvements from MLP-QA to MLP-IQA tend to be lower when facing \IU-decoys. This is also expected as it is difficult to have decoys that are simultaneously both \IU\ and \QU\ --- such answers tend to be the target answers.  Nonetheless, we deem this as a future direction to explore.

\begin{table}
\vskip 3pt
\small
\centering
\tabcolsep 5pt
\begin{tabular}{l|c|c|c|c|c}
Method & Orig & \IU & \QU & \IU+\QU & All \\
\hline
MLP-A   & 31.2 & 39.9 & 45.7 & 31.2 & 27.4 \\
MLP-IA  & 42.0 & 39.8 & 55.1 & 34.1 & 28.7 \\
MLP-QA  & 58.0 & 84.7 & 55.1 & 54.4 & 50.0 \\
MLP-IQA & 64.6 & 85.2 & 65.4 & 63.7 & 58.9 \\
\hline
HieCoAtt$*$ & 63.0 & - & - & 63.7 & - \\
Attntion$*$ & 66.0 & - & - & 66.7 & - \\
\hline
Human  & 88.5$^\dagger$ & - & - & 89.0 & - \\
\hline
Random & 5.6 & 25.0 & 25.0 & 14.3 & 4.2 \\
\end{tabular}
{\\$*$: based on our implementation or modification\\$^\dagger$: taken from \cite{antol2015vqa}}
\vskip -3 pt
\caption{Accuracy (\%) on the validation set in VQA.}
\label{vqa-1}
%\vskip -10pt
\end{table}

\paragraph{Differences across datasets} Contrasting Visual7W to VQA (on the column \IU+\QU), we notice that Visual7W tends to have bigger improvements in general. This is due to the fact that VQA has many questions with ``Yes'' or ``No'' as the targets --- the only valid decoy to the target ``Yes'' is ``No'', and vice versa. As such  decoys are already captured by Orig of VQA (`Yes'' and ``No'' are both top frequently-occurring targets), adding other decoy answers will not make any noticeable improvement. In Supplementary Material, however, we show that once we remove such questions/answers pairs, the degree of improvements increases substantially.

\begin{table}[t]
	\vskip 3pt
	\small
	\centering
	\tabcolsep 5pt
	\begin{tabular}{l|c|c|c}
		Method  & \IU & \QU & \IU+\QU \\
		\hline
		MLP-A    & 29.1 & 36.2 & 19.5 \\
		MLP-IA   & 29.5 & 60.2 & 25.2 \\
		MLP-QA   & 89.3 & 45.6 & 43.9 \\
		MLP-IQA  & 89.2 & 64.3 & 58.5 \\
		\hline
		HieCoAtt$*$  & - & - & 57.5 \\
		Attntion$*$ & -  & - & 60.1\\
		\hline
		Human  & -  & - & 82.5 \\
		\hline
		Random  & 25.0 & 25.0 & 14.3 \\
	\end{tabular}
	{\\$*$: based on our implementation or modification}
	\vskip -3 pt
	\caption{Test accuracy (\%) on qaVG.}
	\label{VG-1}
	\vskip -5pt
\end{table}

\paragraph{Comparison on Visual QA models} As presented in Table~\ref{v7w-1} and \ref{vqa-1}, MLP-IQA is on par with or even outperforms \textbf{Attention} and \textbf{HieCoAtt} on the Orig decoys, showing how the shortcuts make it difficult to compare different models. By eliminating the shortcuts (\ie, on the combined \IU+\QU -decoys), the advantage of using sophisticated models becomes obvious (\textbf{Attention} outperforms MLP-IQA by 3\% in Table \ref{vqa-1}), indicating the importance to design advanced models for achieving human-level performance on Visual QA.
\vspace{10pt}

For completeness, we include the results on the Visual Genome dataset in Table~\ref{VG-1}. This dataset has no ``Orig'' decoys, and we have created a multiple-choice based dataset \textbf{qaVG} from it for the task --- it has over 1 million triplets, the largest dataset on this task to our knowledge. On the combined \IU+\QU -decoys, we again clearly see that machines need to use all the information to succeed.

With \textbf{qaVG}, we also investigate whether it can help improve the multiple-choice performances on the other two datasets. We use the MLP-IQA trained on qaVG with both \IU\ and \QU\ decoys to initialize the models for the Visual7W and VQA datasets. We report the accuracies before and after fine-tuning, together with the best results learned solely on those two datasets. As shown in Table~\ref{t_pretrain}, fine-tuning largely improves the performance, justifying the finding by Fukui et al.~\shortcite{fukui2016multimodal}.

\begin{table}[t]
\vskip 3pt
\small
\centering
\tabcolsep 4.5pt
\begin{tabular}{l|c|c|c|c}
Datasets & Decoys & Best w/o & \multicolumn{2}{c}{qaVG model}\\
\cline{4-5}
& & using qaVG & initial & fine-tuned\\
\hline
\multirow{3}{*}{Visual7W}
& Orig       & 65.7 & 60.5 & \textbf{69.1} \\
& \IU+\QU   & 52.0 & 58.1 & \textbf{58.7} \\
& All & 45.1 & 48.9 & \textbf{51.0} \\
\hline

\multirow{3}{*}{VQA}
& Orig       & 64.6 & 42.2 & \textbf{65.6} \\
& \IU+\QU   & 63.7 & 47.9 & \textbf{64.1} \\
& All & 58.9 & 37.5 & \textbf{59.4} \\
\end{tabular}
\vskip -3 pt
\caption{Using models trained on qaVG to improve Visual7W and VQA (Accuracy in \%).}
\label{t_pretrain}
\vskip -5pt
\end{table}

\subsection{Qualitative Results}
% !TEX root = main.tex
\label{S_a_quality}

\begin{figure*}[t]
	\centering
	\includegraphics[width=0.975\textwidth]{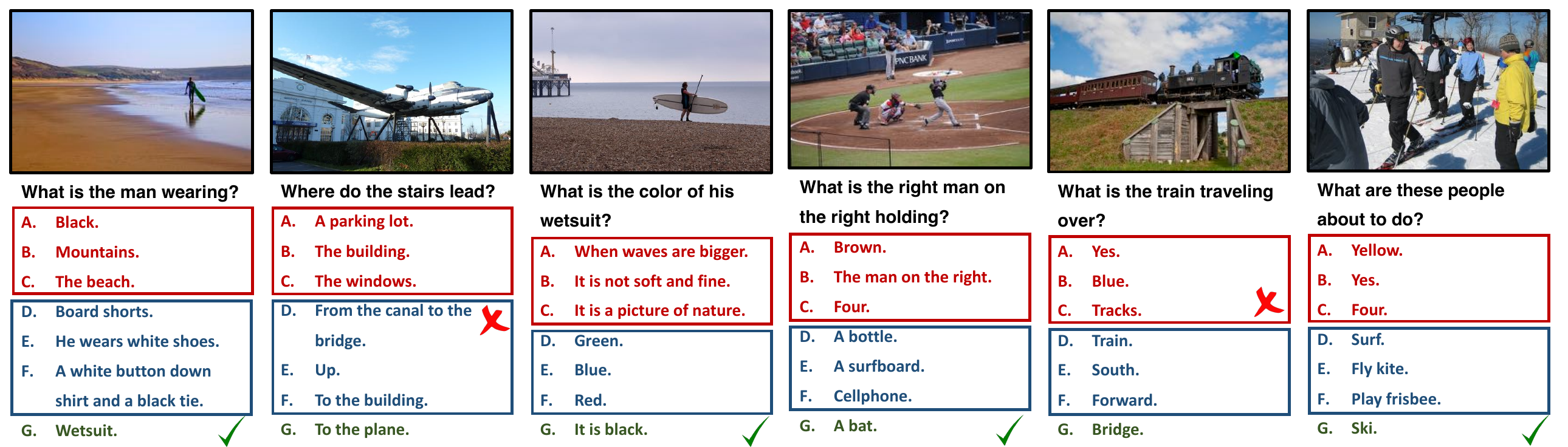}
	\vskip -10 pt
	\caption{Example image-question-target triplets from Visual7W, VQA, and VG, together with our \IU-decoys (A, B, C.) and \QU-decoys (D, E, F). G is the target. Machine's selections are denoted by green ticks (correct) or red crosses (wrong).}
	\label{f_good}
	\vskip -10 pt
\end{figure*}

\begin{figure}[h]
\centering
\includegraphics[width=0.48\textwidth]{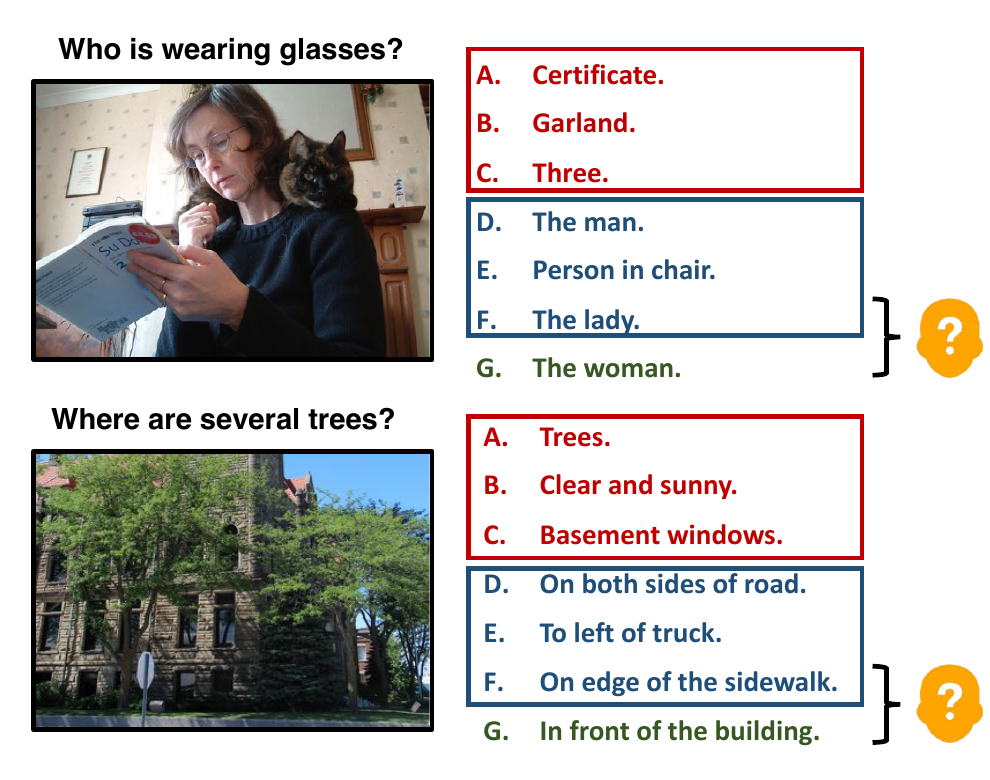}
\vskip -15 pt
\caption{Ambiguous examples by our \IU-decoys (A, B, C) and \QU-decoys (D, E, F). G is the target. Ambiguous decoys F are marked.}
\label{f_bad}
\vskip -10 pt
\end{figure}

In Fig.~\ref{f_good}, we present examples of image-question-target triplets from \textbf{V7W}, \textbf{VQA}, and \textbf{VG}, together with our \IU-decoys (A, B, C) and \QU-decoys (D, E, F). G is the target. The predictions by the corresponding MLP-IQA are also included. Ignoring information from images or questions makes it extremely challenging to answer the triplet correctly, even for humans.

Our automatic procedures do fail at some triplets, resulting in ambiguous decoys to the targets. See Fig.~\ref{f_bad} for examples. We categorized those failure cases into two situations.
\begin{itemize}[leftmargin=2\labelsep]
\item Our filtering steps in Sect.~\ref{S_method} fail, as observed in the top example. The WUP-based similarity relies on the WordNet hierarchy. For some semantically similar words like ``lady'' and ``woman'', the similarity is only 0.632, much lower than that of 0.857 between ``cat'' and ``dog''. This issue can be alleviated by considering alternative semantic measures by \textsc{word2vec} or by those used in \cite{das2016visual,ding2016understanding} for searching similar questions.

\item The question is ambiguous to answer. In the bottom example in Fig.~\ref{f_bad}, both candidates D and F seem valid as a target. Another representative case is when asked about the background of a image. In images that contain sky and mountains in the distance, both terms can be valid.

\end{itemize}

% !TEX root = main.tex
\section{Conclusion}
\label{S_disc}
We perform detailed analysis on existing datasets for multiple-choice Visual QA. We found that the design of decoys can inadvertently provide ``shortcuts'' for machines to exploit to perform well on the task. We describe several principles of constructing good decoys and propose automatic procedures to remedy existing datasets and create new ones. We conduct extensive empirical studies to demonstrate the effectiveness of our methods in creating better Visual QA datasets. The remedied datasets and the newly created ones are released and available at \url{http://www.teds.usc.edu/website_vqa/}.

\section*{Acknowledgments}
This work is partially supported by USC Graduate Fellowship, NSF IIS-1065243, 1451412, 1513966/1632803, 1208500, CCF-1139148, a Google Research Award, an Alfred. P. Sloan Research Fellowship and ARO\# W911NF-12-1-0241 and W911NF-15-1-0484.

\bibliography{vqa}
\bibliographystyle{acl_natbib}

\clearpage
\appendix
\begin{center}
	\textbf{\Large Supplementary Material}
\end{center}

In this Supplementary Material, we provide details omitted in the main text. 
\begin{itemize}[leftmargin=4.5mm]
	\setlength\itemsep{0em}
	\item Sect.~\ref{S_1}: Details on the MLP-based models and the attention-based models (Sect. 3.1 and 5.2 of the main text).
	\item Sect.~\ref{S_2}: WUPS-based similarity for filtering out ambiguous decoys (Sect. 4.1 of the main text).
	\item Sect.~\ref{S_3}: Detailed results on VQA~\cite{antol2015vqa} w/o question-answer (QA) pairs that have Yes/No as the targets (Sect. 5.3 of the main text).
	\item Sect.~\ref{S_4}: Experiments on VQA2 \cite{goyal2016making} and COCOQA \cite{ren2015exploring} (Sect. 5 of the main text).
	\item Sect.~\ref{S_5}: Details on user studies (Sect. 5.2 of the main text).
	\item Sect.~\ref{S_8}: Analysis on different question and answer embeddings (Sect. 5.2 of the main text).
	\item Sect.~\ref{S_7}: Analysis on random decoys (Sect. 5.3 of the main text).
\end{itemize}

% !TEX root = main.tex
\section{Details on the MLP-based models and the attention-based models}
\label{S_1}

As mentioned in the main text, we benchmark the performance of popular Visual QA models on our remedied dataset. Here we provide the details about those models we experimented and its corresponding training configurations. 

\subsection{The simple MLP-based model}
\label{s_concat}
The one hidden-layer MLP model used in our experiments has 8,192 hidden units, exactly following~\cite{jabri2016revisiting}. It contains a batch normalization layer before ReLU, and a dropout layer after ReLU. We set the dropout rate to be 0.5.

\begin{figure}[t!]
	\centering
	\includegraphics[width=0.485\textwidth]{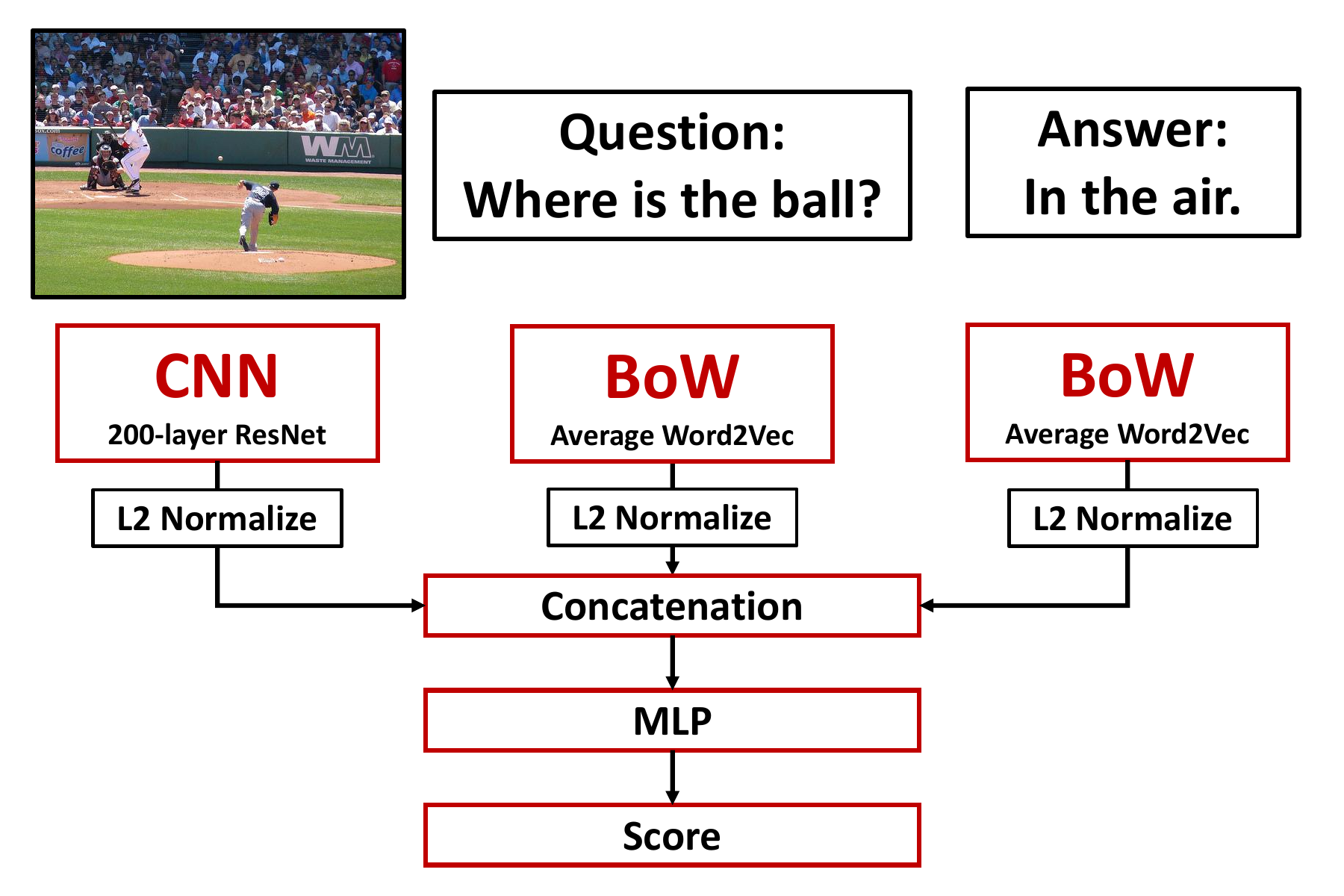}
	\caption{Illustration of MLP-based models.}
	\label{f_mlp}
\end{figure}

The input to the model is the concatenated features of images, questions, and answers, as shown in Fig.~\ref{f_mlp}. We change all characters to lowercases and all integer numbers within $[0, 10]$ to words before computing \textsc{word2vec}. We perform $\ell_2$ normalization to features of each information before concatenation.

\begin{figure*}[t!]
	\small
	\centering
	\tabcolsep 1pt
	\begin{tabular}{cc}
		\includegraphics[width=0.485\textwidth]{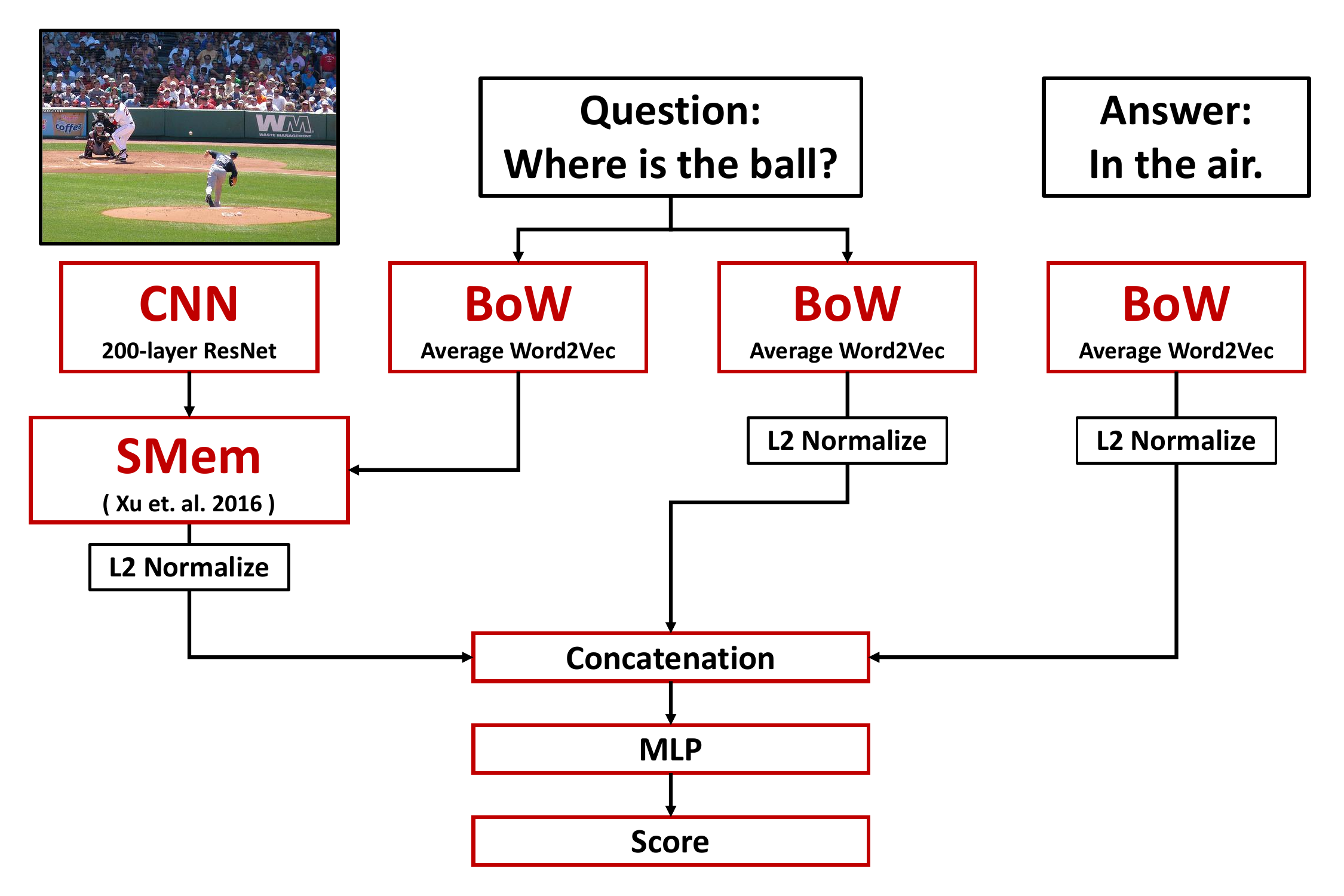} & \includegraphics[width=0.485\textwidth]{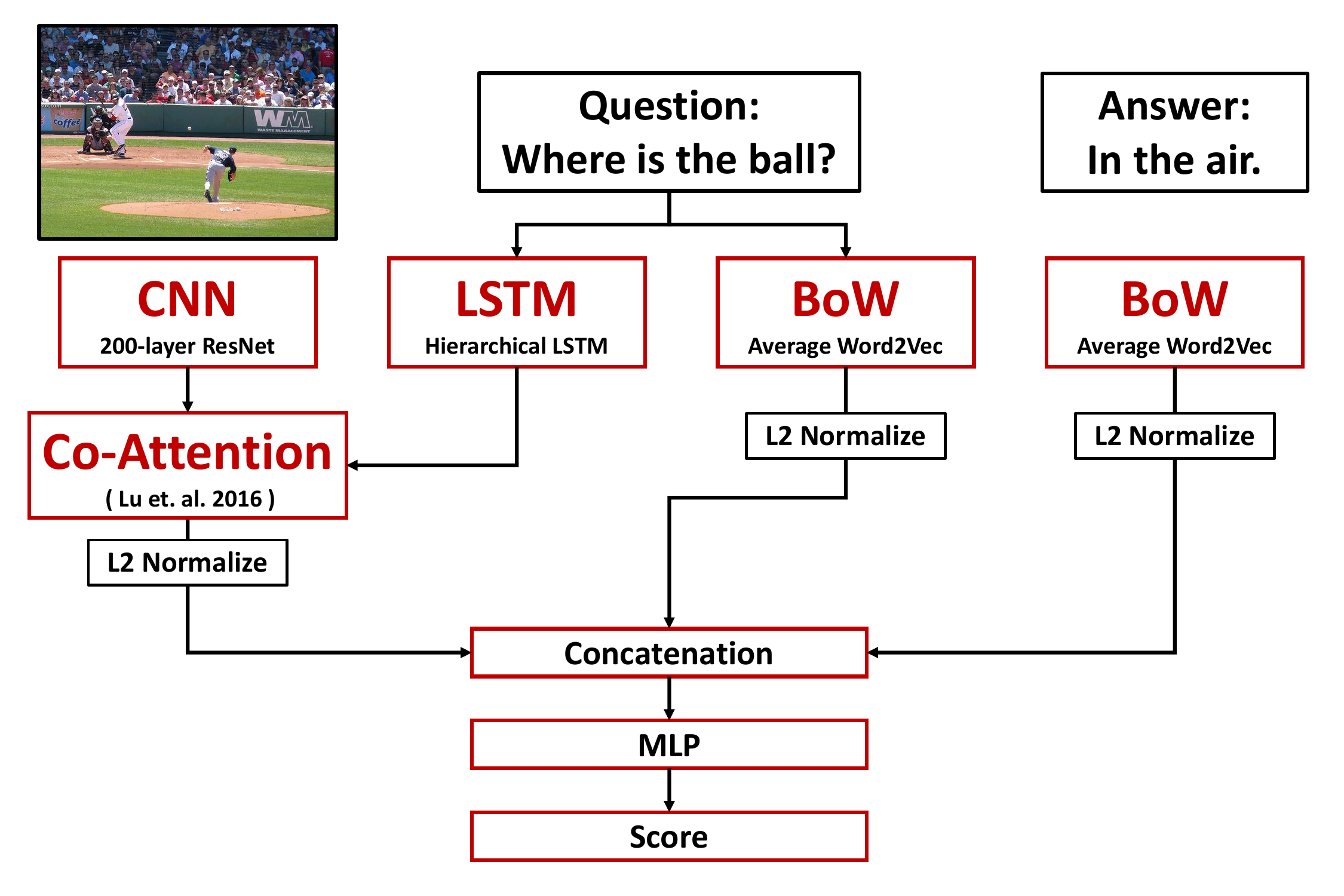} \\
		(a) Attention* (SMem + MLP) & (b) HieCoAtt* (HieCoAtt + MLP) \\
	\end{tabular}
	\caption{Illustration of attention-based Visual QA models.}
	\label{f_att}
\end{figure*}

\subsection{A variant of SMem (Attention*)}
\label{s_bow}
In the main text we experiment with a straightforward attention model similar to the spatial memory network (SMem)~\cite{xu2016ask}, as shown in Fig.~\ref{f_att} (a). Instead of computing the visual attention for each word in the question, we directly compute the visual attention for the entire question using its average \textsc{word2vec} embedding. We then concatenate the resulting visual features with the feature of the question and a candidate answer (in the same way as the MLP-based model in Sect.~\ref{s_concat}) as the input to train a one-hidden-layer MLP for binary classification.

\subsection{A variant of HieCoAtten (HieCoAtten*)}
\label{s_lstm}
Beyond the Attention*, we also experimented HieCoAtt*, a variant of the model proposed by Lu et al.~\shortcite{lu2016hierarchical} (as shown in Fig.~\ref{f_att} (b)). Our model inherits all components in \cite{lu2016hierarchical} that are related to computing the joint multi-modal embedding (from images and questions). To adapt to the multiple-choice setting, we discard the multi-way classifier in the original HieCoAtt and use its penultimate activations as feature for images. Similarly, we then concatenate this together with the features of questions and candidate answers, and input it to a one-hidden-layer MLP, following exact the configuration as Sect.~\ref{s_concat}. 

\subsection{Optimization}

We train all our models using stochastic gradient based optimization method with mini-batch size of 100, momentum of 0.9, and the stepped learning rate policy: the learning rate is divided by 10 after every $M$ mini-batches. We set the initial learning rate to be 0.01 (we further consider 0.001 for the case of fine-tuning in Sect. 5.3 of the main text). For each model, we train with at most 600,000 iterations. We treat $M$ and the number of iterations as hyper-parameters of training. We tune the hyper-parameters on the validation set.

Within each mini-batch, we sample 100 IQA triplets. For each triplet, we randomly choose to use \QU-decoys or \IU-decoys when training on \IU+\QU, or \QU-decoys or \IU-decoys or Orig when training on All. We then take the target and \textbf{3} decoys for each triplet to train the binary classifier (\ie, minimize the logistic loss). Specifically on VQA, which has 17 Orig decoys for a triplet, we randomly choose 3 decoys out of them. That is, 100 triplets in the mini-batch corresponds to 400 examples with binary labels. This procedure is to prevent \emph{unbalanced training}, where machines simply learn to predict the dominant label, as suggested by Jabri et al.~\shortcite{jabri2016revisiting}.  

We note that in all the experiments in the main text, we use the \emph{same type of decoy sets for training and testing}.

\section{WUP-based similarity for filtering out ambiguous decoys}
\label{S_2}

We use the Wu-Palmer (WUP) score~\cite{wu1994verbs}, which characterizes the \emph{word sense} similarity, to filter out ambiguous decoys to the target (correct answer). The WUP score is computed based on the WordNet hierarchy. Essentially, it measures the similarity of two \emph{nodes} (\ie, synsets) in the hierarchy. As a \emph{word} might correspond to multiple nodes, we measure the word similarity as follows:
\begin{align}
WUP(w_1, w_2) = \max_{(n_1, n_2)\in N_1 \times N_2} WUP(n_1, n_2),
\end{align}
where $N_1$ and $N_2$ are the sets of nodes that words $w_1$ and $w_2$ correspond to, respectively. That is, the word similarity is based on the most similar pair of nodes from both words. We consider only the NOUN and ADJ nodes for tractable computation.

Since a candidate answer may contain more than one word (\ie, a word sequence), we compute the similarity between two word sequences $WS_1$ and $WS_2$ as follows 
\begin{align}
\label{e_WUP}
& WUP(WS_1, WS_2) = \nonumber \\
& \max \{\prod_{w_1 \in WS_1} \max_{w_2 \in WS_2} WUP(w_1, w_2), \\
& \hspace{24pt} \prod_{w_2 \in WS_2} \max_{w_1 \in WS_1} WUP(w_1, w_2)\}. \nonumber 
\end{align}
This formulations is highly similar to the one proposed by Malinowski and Fritz et al.~\shortcite{malinowski2014multi} for evaluating open-ended Visual QA. The main difference is that we use ``max'' rather than ``min'' to compute the final score. Note that our purpose of using the WUP score is to filter out ambiguous decoys to the target. For example, we consider ``a cute cat'' to be ambiguous to ``cat''. Using eq. (\ref{e_WUP}) gives a similarity 1, which can not be achieved by taking ``min''.

\subsection{Analysis on the coverage}
Among all the 139,868 IQA triplets in Visual7W, the target answers of 137,557 of them (~98\%) can find corresponding nodes in the WordNet hierarchy, so the scores can be computed. For VQA, the ratio is ~97\%. For qaVG, the ratio is ~99\%.

\section{Detailed results on VQA w/o QA pairs that have Yes/No as the targets}
\label{S_3}
As mentioned in Sect. 5.3 of the main text, the validation set of VQA contains 45,478 QA pairs (out of totally 12,1512 pairs) that have Yes or No as the correct answers. The only reasonable decoy to Yes is No, and vice versa --- any other decoy could be easily recognized in principle. Since both of them are among top 10 frequently-occurring answers, they are already included in the Orig decoys --- our \IU-decoys and \QU-decoys can hardly make noticeable improvement. We thus remove all those pairs (denoted as Yes/No QA pairs) to investigate the improvement on the remaining pairs, for which having multiple choices makes sense. We denote the subset of VQA as VQA$^-$ (we remove Yes/No pairs in both training and validation set).

\begin{table}[t]
	\centering
	\small
	\tabcolsep 5pt
	\begin{tabular}{l|c|c|c|c|c}
		Method & Orig & \IU & \QU & \IU+\QU & All \\
		\hline
		MLP-A   & 28.8 & 42.9 & 34.5 & 23.6 & 15.8 \\
		MLP-IA  & 43.0 & 44.8 & 53.2 & 35.5 & 28.5 \\
		MLP-QA  & 45.8 & 80.7 & 39.3 & 38.2 & 31.9 \\
		MLP-IQA & 55.6 & 81.8 & 56.6 & 53.7 & 46.5 \\
		\hline
		HieCoAtt$*$ & 54.8 & - & - & 55.6 & - \\
		Attention$*$ & 58.5 & - & - & 58.6 & - \\ 
		\hline
		Human-IQA & - & - & - & 85.5 & - \\
		\hline
		Random & 5.6 & 25.0 & 25.0 & 14.3 & 4.2 \\
	\end{tabular}
	{\\$*$: based on our implementation or modification}
	\caption{Accuracy (\%) on VQA$^-$-2014val, which contains 76,034 triplets.}
	\label{vqa-2}
	\vskip -5pt
\end{table}

We conduct the same experiments as in Sect. 5.3 of the main text on VQA$^-$. Table~\ref{vqa-2} summarizes the machines' as well as humans' results. Compared to Table 4 of the main text, most of the results drop, which is expected as those removed Yes/No pairs are considered simpler and easier ones --- their \emph{effective} random chance is 50\%. The exception is for the MLP-IA models, which performs roughly the same or even better on VQA$^-$, suggesting that Yes/No pairs are somehow difficult to MLP-IA. This, however, makes sense since without the questions (e.g., those start with ``Is there a ...'' or ``Does the person ...''), a machine cannot directly tell if the correct answer falls into Yes or No, or others.

We see that on VQA$^-$, the improvement by our \IU-decoys and \QU-decoys becomes significant. The gain brought by images on \QU\ (from 39.3\% to 56.6\%) is much larger than that on Orig (from 45.8\% to 55.6\%). Similarly, the gain brought by questions on \IU\ (from 44.8\% to 81.8\%) is much larger than that on Orig (from 43.0\% to 55.6\%). After combining \IU-decoys and \QU-decoys as in \IU+\QU\ and All, the improvement by either including images to MLP-QA or including questions to MLP-IA is noticeable higher than that on Orig. Moreover, even with only 6 decoys, the performance by MLP-A on \IU+\QU\ is already lower than that on Orig, which has 17 decoys, demonstrating the effectiveness of our decoys in preventing machines from overfitting to the incidental statistics. These observations together demonstrate how our proposed ways for creating decoys improve the quality of multiple-choice Visual QA datasets.

\section{More experimental results on VQA2 and COCOQA}
\label{S_4}

\subsection{Dataset descriptions}

\paragraph{COCOQA~\cite{ren2015exploring} }
This dataset contains in total 117,684 auto-generated IQT triplets with no decoy answers. Therefore, we create decoys using our proposed approach and follow the original data split, leading to a training set and a testing set with 78,736 IQA triplets and 38,948 IQA triplets, respectfully.

\paragraph{VQA2~\cite{goyal2016making}}
VQA2 is a successive dataset of VQA, which pairs each IQT triplet with a complementary one to reduce the correlation between questions and answers. There are 443,757 training IQT triplets and 214,354 validation IQT triplets, with no decoys. We generate decoys using our approach and follow the original data split to organize the data. We do not consider the test split as it does not indicate the targets (correct answers).

\subsection{Experimental results}

For both datasets, we conduct the same experiments as in Sect. 5.3 of the main text using the MLP-based models. As shown in Table~\ref{COCOQA-1}, we clearly see that with only answers being visible to the model (MLP-A), the performance is close to random (on the column of \IU+\QU -decoys), and far from observing all three sources of information (MLP-IQA). Meanwhile, models that can observe either images and answers (MLP-IA) or questions and answers (MLP-QA) fail to predict as good as the model that observe all three sources of information. Results in Table~\ref{VQA2-1} also shows a similar trend. These empirical observations meet our expectation and again verify the effectiveness of our proposed methods for creating decoys.

\begin{table}[t]
	\small
	\centering
	\tabcolsep 5pt
	\begin{tabular}{l|c|c|c}
		Method  & \IU & \QU & \IU+\QU \\
		\hline
		MLP-A    &  70.3 & 31.7 & 26.6 \\
		MLP-IA   & 73.4 & 73.3 & 60.7 \\
		MLP-QA   &  91.5 & 52.5 & 51.4 \\
		MLP-IQA  & 93.1 & 78.3 & 75.9 \\
		\hline
		Random  & 25.0 & 25.0 & 14.3 \\
	\end{tabular}
	\caption{Test accuracy (\%) on COCOQA.}
	\label{COCOQA-1}
	%\vskip -5pt
\end{table}

\begin{table}[t]
\small
\centering
\tabcolsep 5pt
\begin{tabular}{l|c|c|c}
Method  & \IU & \QU & \IU+\QU \\
\hline
MLP-A    &  37.7 & 41.9 & 27.7 \\
MLP-IA   &  37.9 & 54.4 & 30.5 \\
MLP-QA   & 84.2 & 48.3 & 48.1 \\
MLP-IQA  & 86.3 & 63.0 &  61.1\\
\hline
Random  & 25.0 & 25.0 & 14.3 \\
\end{tabular}
\caption{Test accuracy (\%) on VQA2-2017val.}
\label{VQA2-1}
\vskip -5pt
\end{table}

Besides, we also perform a more in-depth experiment on VQA2, removing triplets with Yes/No as the target. We name this subset as VQA2$^-$. Table~\ref{VQA2-2} shows the experimental results on VQA2$^-$. Comparing to Table~\ref{VQA2-1}, we see that the overall performance for each model decreases as the dataset becomes more challenging on average. Specifically, the model that observes question and answer on VQA2$^-$ performs much worse than that on VQA2 (37.2\% vs. 48.1\%). 

\begin{table}[t]
	\small
	\centering
	\tabcolsep 5pt
	\begin{tabular}{l|c|c|c}
		Method  & \IU & \QU & \IU+\QU \\
		\hline
		MLP-A   & 39.8 & 33.7 &  21.3 \\
		MLP-IA   &  40.3 &  53.0 & 31.0 \\
		MLP-QA   & 84.8 & 37.6 & 37.2 \\
		MLP-IQA  &  85.9 & 56.1 &  53.8\\
		\hline
		Random  & 25.0 & 25.0 & 14.3 \\
	\end{tabular}
	\caption{Test accuracy (\%) on VQA2$^-$-2017val, which contains 134,813 triplets.}
	\label{VQA2-2}
	\vskip -5pt
\end{table} 

\section{Details on user studies}
\label{S_5}

\begin{figure*}
	\centering
	\includegraphics[width=0.95\textwidth]{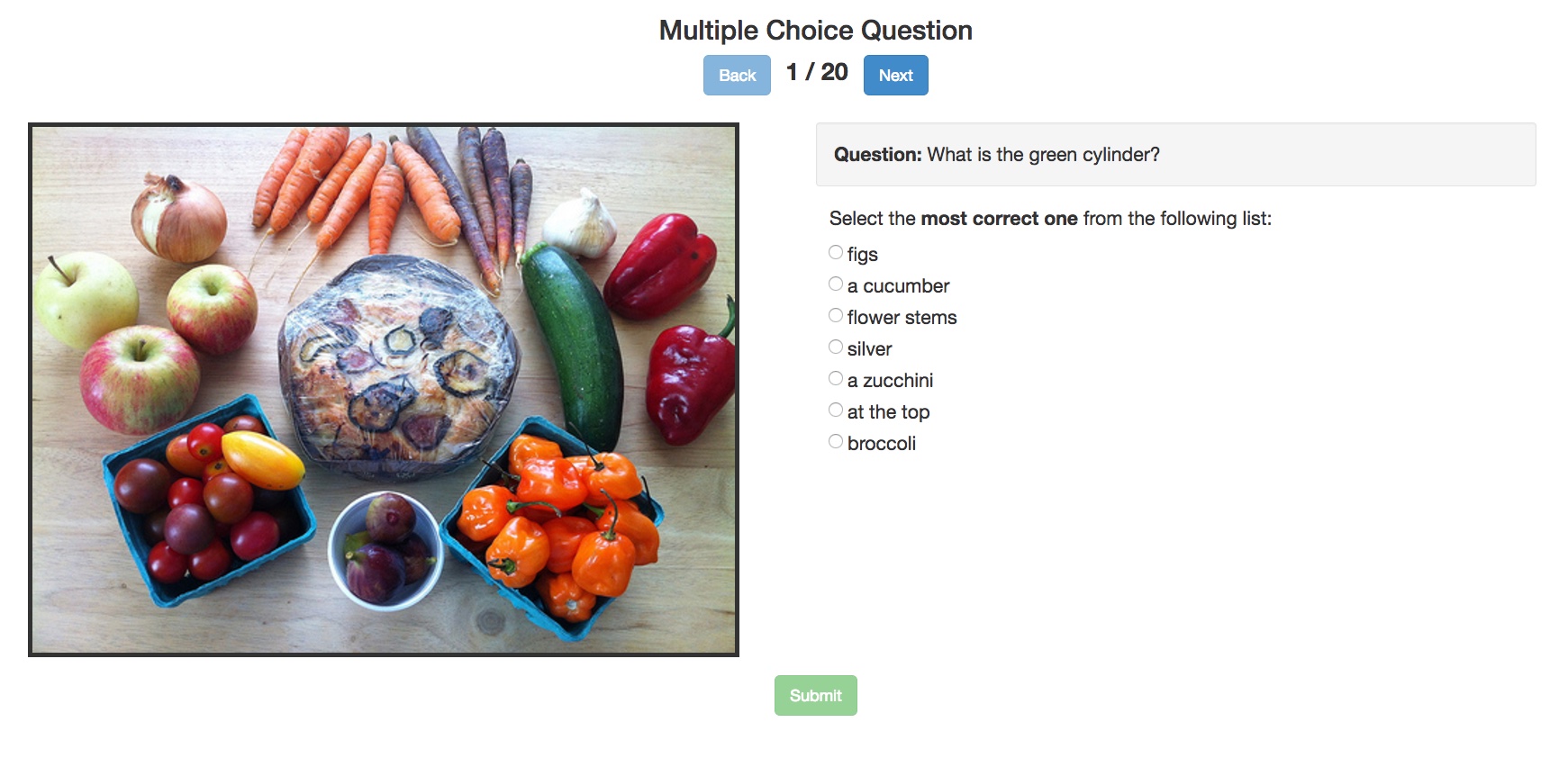}
	\caption{User interface for human evaluation on Visual7W (\IU-decoys+\QU-decoys).}
	\label{f_ui_iqa}
	\vskip -5pt
\end{figure*}

As mentioned in Sect.~5.2 of the main text, we provide details on user studies. Fig.~\ref{f_ui_iqa} shows our user interface. We perform the studies using Amazon Mechanic Turk (AMT) on Visual7W~\cite{zhu2016Visual7W}, VQA~\cite{antol2015vqa} and Visual Genome (VG)~\cite{krishna2016vg}. We mainly evaluate on our \IU-decoys and \QU-decoys (combined together). 

For each dataset, we randomly sample 1,000 image-question-target triplets together with the corresponding \IU-decoys and \QU-decoys to evaluate human performance. For each of these triplets, three workers are assigned to select the most correct candidate answer according to the image and the question. We compute the average accuracy of these workers and report them in Table~3, 4 and 5 of the main text and Table~\ref{vqa-2}.  

\begin{figure*}
	\centering
	\includegraphics[width=0.95\textwidth]{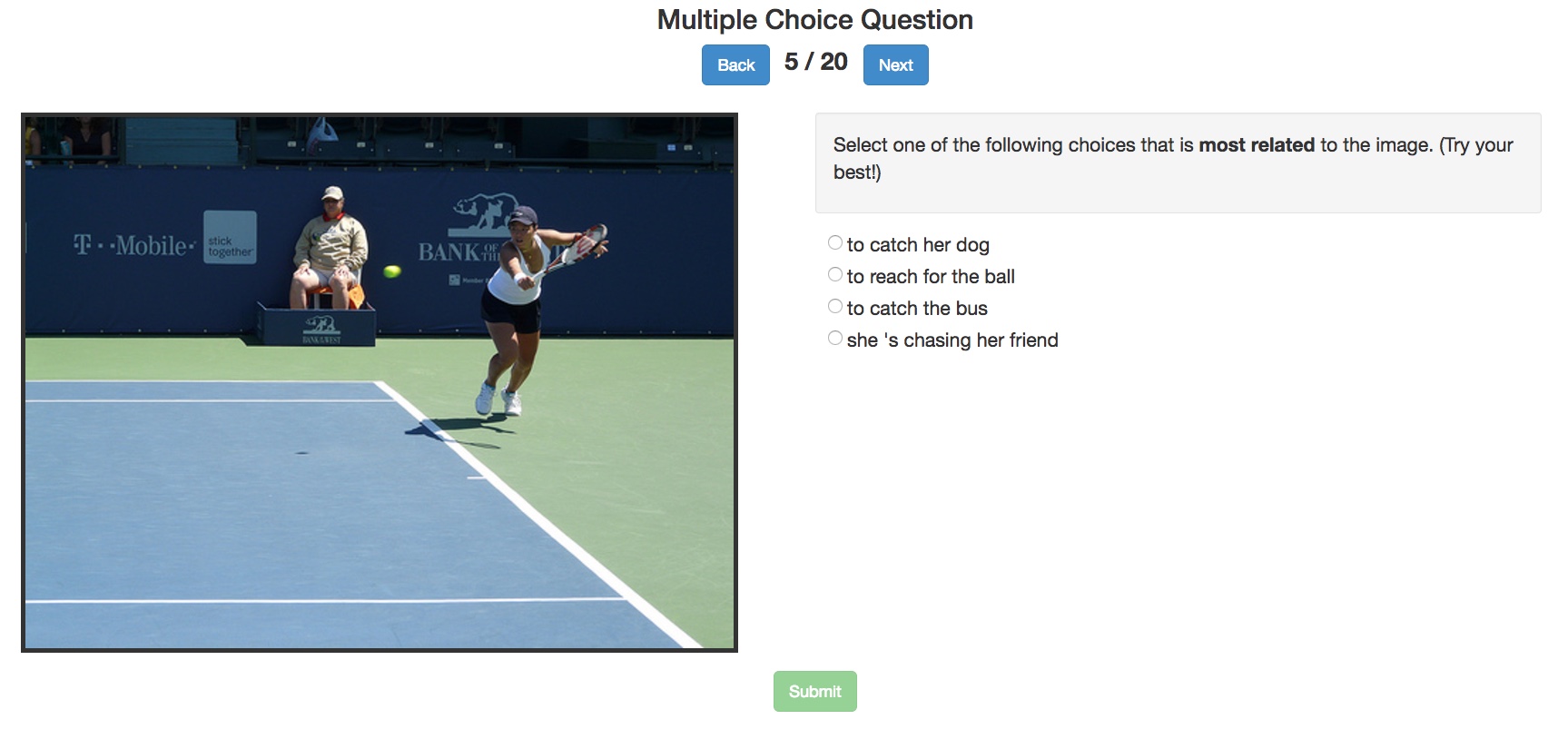}
	\caption{User interface for human evaluation on Visual7W (Orig decoys), where questions are blocked.}
	\label{f_ui_ia}
\end{figure*}

\begin{figure*}
	\centering
	\includegraphics[width=0.95\textwidth]{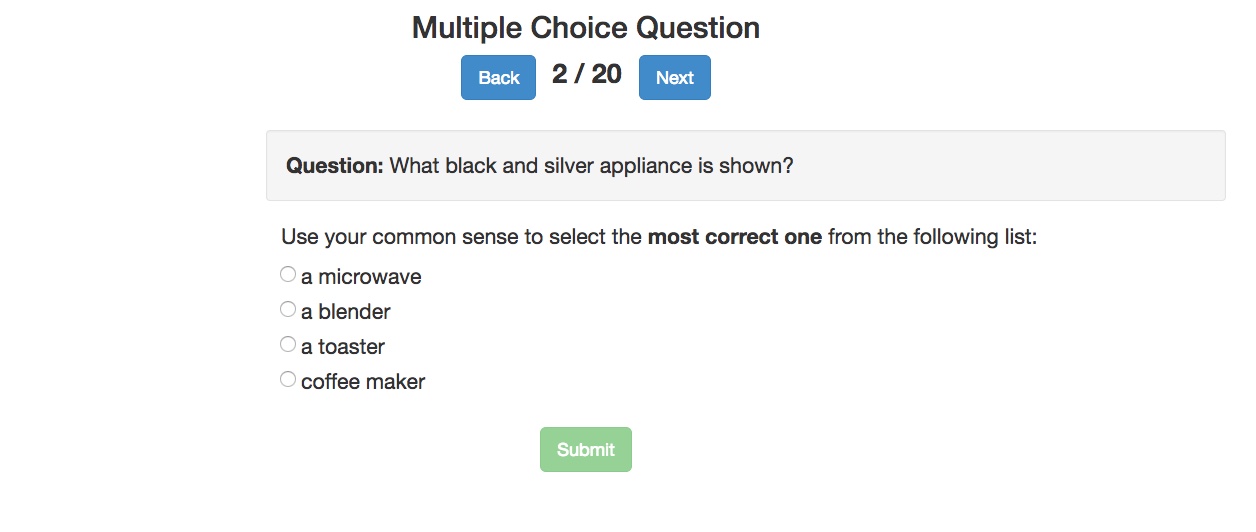}
	\caption{User interface for human evaluation on Visual7W (Orig decoys), where images are not blocked.}
	\label{f_ui_qa}
\end{figure*}

We also conduct human evaluation using the Orig decoys of Visual7W so as to investigate the difference between human-generated and automatically generated decoys. We also study how humans will perform given only partial information (\ie, images + candidate answers or questions + candidate answers), again using the Orig decoys of Visual7W. The corresponding interfaces are shown in Fig.~\ref{f_ui_ia} and~\ref{f_ui_qa}. For these studies, we use the same set of 1,000 triplets used to evaluate our created decoys for fair comparison. We make sure that no worker works on the same triplet across the four studies on Visual7W. Results are reported in Table 1 of the main text.

In summary, 169 workers are involved in our studies. The total cost is \$215 --- the rate for every 20 triplets is \$0.25. On our \IU-decoys and \QU-decoys, humans achieve 84.1\%, 89.0\%, and 82.5\% on Visual7W, VQA, and VG, respectively. Compared to the human performance on the Orig decoys that involve human effort in creation (\ie, 88.4\% on Visual7W, and 88.5\% on VQA as reported in~\cite{antol2015vqa}), these results suggest that the ways we create the decoys and the filtering steps mentioned in Sect. 4.2 lead to high-quality datasets with limited ambiguity. 

\section{Analysis on different question and answer embeddings}
\label{S_8}

\begin{table}
	\small
	\centering
	\tabcolsep 5pt
	\begin{tabular}{l|c|c|c}
		Method  & \textsc{GloVe} & Translation & \textsc{word2vec} \\
		\hline
		MLP-A   & 18.0 & 18.0 & 17.7 \\
		MLP-IA  & 23.6 & 23.2 & 23.6 \\
		MLP-QA  & 38.1 & 38.3 & 37.8 \\
		MLP-IQA & 52.5 & 51.4 & 52.0 \\
		\hline
		Random & 14.3 & 14.3  & 14.3 \\
	\end{tabular}
	\caption{Test accuracy (\%) on Visual7W, comparing different embeddings for questions and answers. The results are reported for the \IU+\QU-decoys.}
	\label{s-v7w-2}
	%\vskip -5pt
\end{table}

We consider \textsc{GloVe}~\cite{pennington2014glove} and the embedding learned from translation~\cite{McCann2017LearnedIT} on both question and answer embeddings. The results on Visual7W (IoU + QoU, compared to Table 3 of the main text that uses \textsc{word2vec}) are in Table~\ref{s-v7w-2}. We do not observe significant difference among different embeddings, which is likely due to that both the questions and answers are short (averagely 7 words for questions and 2 for answers).

\section{Analysis on random decoys}
\label{S_7}

\begin{table}
	\small
	\centering
	\tabcolsep 5pt
	\begin{tabular}{l|c|c|c}
		Method  & (A) & (B) & All \\
		\hline
		MLP-A   & 39.6 & 11.6 & 15.6 \\
		MLP-IA  & 53.4 & 40.3 & 22.2 \\
		MLP-QA  & 52.3 & 50.3 & 31.9 \\
		MLP-IQA & 61.5 & 60.2 & 45.1 \\
		\hline
		Random & 10.0 & 10.0  & 10.0 \\
	\end{tabular}
	\caption{Test accuracy (\%) on Visual7W, comparing different random decoy strategies to our methods: (A) Orig + uniformly random decoys from unique correct answers, (B) Orig + weighted random decoys w.r.t. their frequencies, and All (Orig+\IU+\QU).}
	\label{s-v7w-1}
	\vskip -5pt
\end{table}

We conduct the analysis on sampling random decoys, instead of our \IU-decoys and \QU-decoys, on Visual7W. We collect 6 additional random decoys for each \textbf{Orig} IQA triplet so the answer set will contain 10 candidates, the same as \textbf{All} in Table 3 of the main text. We consider two strategies: (A) uniformly random decoys from unique correct answers, and (B) weighted random decoys w.r.t. their frequencies. The results are in Table~\ref{s-v7w-1}. We see that different random strategies lead to drastically different results. Moreover, compared to the \textbf{All} column in Table 3 of the main text, we see that our methods lead to a larger relative gap between MLP-IQA to MLP-IA and MLP-QA than both random strategies, demonstrating the effectiveness of our methods in creating decoys.

\end{document}